\documentclass[10pt,letterpaper]{article}
\usepackage[shownumpages,citingstyle=numbers,bibliostyle=plainnat,bibfile=refs,boxrule=0.7pt]{brainlab}
\usepackage{shortcuts}
\usepackage{needspace}
\usepackage{placeins}

\hypersetup{urlcolor=braincolor}
\renewcommand{\brainprintbibliography}{%
  \clearpage
  \begingroup\small\setlength{\bibsep}{2pt plus 1pt}
  \bibliographystyle{plainnat}
  \bibliography{refs}%
  \endgroup
}

\definecolor{cteal}{HTML}{2B6E73}
\definecolor{crust}{HTML}{BF572F}
\definecolor{cgold}{HTML}{DEA83E}
\definecolor{csand}{HTML}{E4EEFA}
\definecolor{cink}{HTML}{2D3033}
\definecolor{cgrey}{HTML}{8A867E}
\definecolor{methodblue}{HTML}{2B68A0}
\definecolor{methodrust}{HTML}{A44D32}
\newcommand{\best}[1]{\mathbf{#1}}
\newcommand{\second}[1]{\underline{#1}}
\newcommand{\yes}{\ensuremath{\textcolor{cteal}{\checkmark}}}
\newcommand{\no}{\ensuremath{\textcolor{crust}{\times}}}

\makeatletter
\renewcommand{\subsection}{%
  \@startsection{subsection}{2}{\z@}%
  {-2mm}{0.7mm}{\normalfont\normalsize\bfseries}}
\renewcommand{\paragraph}{%
  \@startsection{paragraph}{4}{\z@}%
  {-2mm}{-2mm}{\normalfont\normalsize\bfseries}}
\makeatother

\providecommand{\method}{\textsc{MaskAhead}\xspace}
\providecommand{\qmethod}{\textsc{Q-MaskAhead}\xspace}

\setbrainmeta{
  title={Mask-Guided KV Cache Eviction\\[2pt]in Block Diffusion Language Models},
  authors={%
\normalfont\mdseries\normalsize\setlength{\tabcolsep}{3pt}%
\begin{tabular*}{\linewidth}{@{\extracolsep{\fill}}ccc@{}}
\parbox[t]{0.322\linewidth}{\centering
  \textbf{Gleb Molodtsov}\textsuperscript{\ensuremath{\dagger}}\\
  {\fontsize{9}{11}\selectfont BRAIn Lab}} &
\parbox[t]{0.322\linewidth}{\centering
  \textbf{Ekaterina Alimaskina}\\
  {\fontsize{9}{11}\selectfont BRAIn Lab\\Yandex Research}} &
\parbox[t]{0.322\linewidth}{\centering
  \textbf{Evgeny Uskov}\\
  {\fontsize{9}{11}\selectfont Yandex School of Data Analysis}}
\\[3.2em]
\multicolumn{3}{c}{%
  \begin{tabular}{@{}cc@{}}
  \parbox[t]{0.38\linewidth}{\centering
    \textbf{Artur Zagitov}\\
    {\fontsize{9}{11}\selectfont BRAIn Lab}} &
  \parbox[t]{0.38\linewidth}{\centering
    \textbf{Aleksandr Beznosikov}\\
    {\fontsize{9}{11}\selectfont BRAIn Lab}}
  \end{tabular}%
}
\end{tabular*}%
  },
  affiliations={{\small\textsuperscript{\ensuremath{\dagger}}Correspondence: \href{mailto:gleb.molodcov@gmail.com}{\texttt{gleb.molodcov@gmail.com}}}},
  code = {\url{https://github.com/brain-lab-research/MaskAhead}},  abstract={Block diffusion language models keep a large key-value (KV) cache throughout generation and attend to it at every denoising step, limiting both memory capacity and generation speed.
Reducing these costs requires deciding which past tokens to use for denoising the current block (selection) and which to keep in memory for future blocks (eviction).
We propose \method, a training-free method that solves both tasks with a single mask-query-based ranking mechanism.
Current-block masks guide selection, while probes of upcoming masked blocks guide eviction.
Both rank KV entries by their estimated contribution to the attention output.
Our quantized variant, \qmethod, computes selection and attention directly from low-bit KV, largely preserving the selected entries.
Experiments on Fast-dLLM-v2, DreamReasoner, and LLaDA2.0-mini cover long-generation reasoning, long-prompt question answering, and needle-in-a-haystack retrieval.
On long-prompt QA, \method reduces KV memory by $9.5\times$ on average with a 1.2-point mean F1 loss relative to dense inference.
\qmethod increases the reduction to $20.1\times$ with a 2.3-point mean F1 loss.
In a batch-32 systems profile, \method achieves $1.23\times$ end-to-end and $1.68\times$ decode-stage speedups over dense inference.}
}

\begin{document}
\begin{mainpart}

\section{Introduction}
\label{sec:intro}

Diffusion language models (dLLMs) generate text by repeatedly denoising masked positions, replacing the token-by-token autoregressive loop with parallel decoding~\citep{lou2024discrete,sahoo2024simple,nie2025llada,ye2025dream}. This parallelism accelerates generation, but over a long decoding horizon it weakens dependencies among jointly predicted tokens \citep{wu2025fastdllm,kang2026parallelbench}. Current systems therefore use block diffusion, generating blocks in sequence while denoising the tokens within blocks in parallel~\citep{arriola2025block,wu2025fastdllmv2,bie2025llada2}.

Block diffusion restores a causal prefix and with it a key-value (KV) cache: after a block is finalized, later blocks reuse its keys and values without recomputation. As the prompt and response grow, every new block retains more KV entries, and every denoising step reads a longer prefix. Long-context inference pays twice for the same history: in resident cache memory and in repeated cache reads.

Recent methods target the second cost. They use mask-token queries to predict the attention distribution of subsequent denoising steps and read only the most relevant prefix entries~\citep{kwon2026mage,kwon2026herald}. This reduces cache reads but retains the full history. Reducing memory instead requires irreversible eviction: deciding which entries future blocks will not need.

Block diffusion permits a query probe before future blocks are generated: masked positions expose model-derived queries conditioned on the known prefix. We use these queries as a forecast for eviction, while current-block queries select temporary reads. The forecast cannot include tokens that intermediate blocks will later finalize. \method tests whether this forward-looking evidence helps most where retention is irreversible and storage is tight.

\section{Related Work}
\label{sec:related}

\paragraph{Sparse reads in block diffusion.}
An exact prefix cache makes block diffusion practical, but its full contents are read repeatedly during denoising. SparseD~\citep{wang2026sparsed} exploits attention patterns shared across steps, while FlashBlock~\citep{chen2026flashblock} reuses stable attention outputs outside the active block. MAGE~\citep{kwon2026mage} instead uses attention from the fully masked first step to select one prefix subset for the remaining denoising steps. These methods reduce cache reads across denoising steps but keep the full prefix resident on the GPU, leaving its memory unchanged.

The attention-trace analysis in \Cref{fig:attention-reuse} tests how long a read set remains informative. We use it as a mechanistic diagnostic, not as evidence that a future-mask probe beats alternative query sources.

HERALD~\citep{kwon2026herald} offloads the full cache to host memory and retrieves a selected subset for each block. It uses one representative mask query to reduce the cost of selection. Offloading lowers GPU-resident memory but retains the full cache in host memory and incurs retrieval and transfer costs. \method instead physically removes entries, enforcing a storage budget after each eviction. GroupKV~\citep{wang2026groupkv} extends block-dLLM retrieval with hierarchical group selection and predictive prefetching across cache tiers.

\paragraph{Eviction in diffusion models.}
Eviction has been studied for full-attention dLLMs, which revisit the entire prompt at every denoising step. Sparse-dLLM~\citep{song2026sparsedllm} chooses entries using saliency that persists across steps, while MaskKV~\citep{huang2025maskkv} ranks prompt entries by mask-query attention and assigns budgets per layer and head. Block diffusion exposes masks only in the active block and continually appends finalized output to an exact cache. This requires recurrent, forward-looking eviction. We test whether future masks provide a robust look-ahead signal and turn it into a practical strategy.

\paragraph{Estimating what future tokens will need in AR generation.}
AR eviction methods must make retention decisions without the queries that future tokens will form. StreamingLLM~\citep{xiao2023streamingllm} protects attention sinks and recent tokens, while H2O~\citep{zhang2023h2o} accumulates past attention to preserve heavy hitters. Our eviction policy protects initial and recent positions, and ranks the remaining entries with future-mask attention.

LookaheadKV~\citep{ahn2026lookaheadkv} learns soft lookahead tokens and adapters to estimate future token importance. It requires additional data and model-specific optimization. Training-free methods instead approximate the missing queries: Expected Attention~\citep{devoto2025expected} extrapolates their distribution from past hidden states, while LAQ~\citep{wang2025laq} and DapQ~\citep{tian2026dapq} construct positional surrogates. CAOTE~\citep{goel2025caote} ranks entries by attention-output sensitivity, combining attention weights with value vectors. Its H2O instantiation uses accumulated past attention. \method uses value-aware scores with two query horizons: current-block masks for temporary reads and upcoming masked blocks for persistent retention. A no-write probe supplies the future queries, and recurrent compaction enforces the storage budget.

\begin{table*}[!b]
\centering
\caption{Native capabilities of the cited methods. Sparse read selects a temporary attention subset; training-free eviction permanently removes KV entries; quantization stores KV at reduced precision.}
\label{tab:literature-map}
\scriptsize\setlength{\tabcolsep}{5pt}\renewcommand{\arraystretch}{1.02}
\begin{tabular}{@{}lccc@{}}
\toprule
Method & Sparse read & Eviction & KV quantization \\
\midrule
Dense bf16 & \no & \no & \no \\
HERALD~\citep{kwon2026herald} & \yes & \no\textsuperscript{*} & \no \\MAGE~\citep{kwon2026mage} & \yes & \no & \no \\
H2O~\citep{zhang2023h2o} & \no & \yes & \no \\
H2O + CAOTE~\citep{zhang2023h2o,goel2025caote} & \no & \yes & \no \\
Expected Attention~\citep{devoto2025expected} & \no & \yes & \no \\
DapQ~\citep{tian2026dapq} & \no & \yes & \no \\
\rowcolor{csand!70}
\method & \yes & \yes & \no \\
\rowcolor{csand!35}
\qmethod & \yes & \yes & \yes \\
\bottomrule
\end{tabular}
\par\smallskip
\noindent\rule{\linewidth}{0.4pt}
\par\smallskip
\begin{minipage}{\linewidth}
\footnotesize
\textsuperscript{*}HERALD offloads the full logical cache to host memory rather than physically evicting entries.
\end{minipage}
\end{table*}

\paragraph{Precision as another storage decision.}
For diffusion models, the DART NPU study introduces block-adaptive online smoothing for KV quantization and evaluates four-bit cache formats under blocked inference~\citep{lou2026dart}. This establishes four-bit KV quantization as prior work in diffusion inference. DART quantizes the full cache representation; it does not combine precision reduction with per-token selection and recurrent eviction. In AR models, KIVI~\citep{liu2024kivi} and KVQuant~\citep{hooper2024kvquant} reduce cache precision, while mixed-precision schemes jointly allocate entries and bits~\citep{yang2024mikv,zhang2025moretokens,zhang2026rdkv}. Q-Hitter~\citep{zhang2024qhitter} shows that quantization can alter importance ranking and develops a quantization-aware selector. We test a complementary setting: a physically compacted cache whose four-bit keys and values are used for both ranking and executed attention, together with recurring mask-query-guided eviction.

\Cref{tab:literature-map} makes the design distinction explicit: existing comparisons cover sparse reads or eviction, while \method combines them and \qmethod adds reduced-precision storage.

\paragraph{Contributions.} Our main contributions can be summarized as follows:
\begin{itemize}[leftmargin=1.4em, itemsep=1pt, topsep=2pt]
\item \textbf{Two decisions, two query horizons.} We distinguish temporary reads from irreversible retention and identify the query evidence each decision requires.
\item \textbf{Future masks guide permanent retention.} \method combines current-block read selection with future-mask probes and value-aware eviction, physically enforcing a persistent KV budget.
\item \textbf{Four-bit storage preserves selection.} We disentangle selection, eviction, and precision. The ranking diagnostic finds 97.44\% read-set and 97.65\% retention-set overlap between four-bit and bf16 storage, supporting \qmethod's compact persistent cache without a bf16 shadow copy.
\item \textbf{Validation across three models and distinct workloads.} Experiments on Fast-dLLM-v2-7B, DreamReasoner-8B, and LLaDA 2.0 cover long-generation reasoning, long-prompt QA, and needle-in-a-haystack (NIAH) passage retrieval. \method achieves the highest F1 among evicting methods in all six main QA settings, including one tie, while reducing KV memory by $9.5\times$ on average.
\end{itemize}

\section{Problem Setup: Read and Retention Decisions}
\label{sec:setup}

\subsection{Two decisions over one prefix}
Block diffusion finalizes one block before decoding the next.
Let $\mathcal I_{t,\ell,h}$ be the live prefix entries at block $t$ for layer $\ell$ and KV head $h$.
During block $t$, a read set $R_{t,\ell,h}\subseteq\mathcal I_{t,\ell,h}$ supplies prefix attention for the remaining denoising steps.
Entries outside $R_{t,\ell,h}$ stay in storage and can be read by a later block.
At an eviction checkpoint, a retained set $S_{t,\ell,h}\subseteq\mathcal I_{t,\ell,h}$ replaces the persistent cache.
An entry outside $S_{t,\ell,h}$ is unavailable until it is recomputed, which the inference procedure does not do.
The read budget $k_t$ and retention budget $C_t$ therefore constrain different resources.

For a query $q$ and prefix set $A$, let $\boldsymbol o(q;A)$ be the prefix-attention output obtained after normalizing attention over $A$.
If $\mathcal Q_t^{\rm later}$ contains later-step queries of block $t$ and $\mathcal Q_j^{\rm realized}$ contains the queries eventually realized in block $j$, the two ideal decisions minimize different errors:
\begin{equation}
\begin{aligned}
\mathcal L_{\rm read}(R_t)
&=\sum\nolimits_{q\in\mathcal Q_t^{\rm later}}
\|\boldsymbol o(q;\mathcal I_t)-\boldsymbol o(q;R_t)\|_2,
& |R_t|&\leq k_t,\\
\mathcal L_{\rm retain}(S_t)
&=\sum\nolimits_{j=t+1}^{t+E}\sum\nolimits_{q\in\mathcal Q_j^{\rm realized}}
\|\boldsymbol o(q;\mathcal I_t)-\boldsymbol o(q;S_t)\|_2,
& |S_t|&\leq C_t.
\end{aligned}
\label{eq:two-objectives}
\end{equation}
The equations describe the decision horizons, not objectives that our implementation optimizes exactly.
In particular, $\mathcal Q_j^{\rm realized}$ is unknown when $S_t$ must be chosen.
The reference set $\mathcal I_t$ excludes KV appended after the checkpoint, so the retention objective isolates the effect of deleting currently live entries. At block $t$, the temporary read buffer is gathered before eviction. The retained set must serve later blocks and the first step at the next checkpoint.

\begin{figure}[t]
\centering
\begin{minipage}[t]{0.47\linewidth}
\centering
\includegraphics[width=\linewidth]{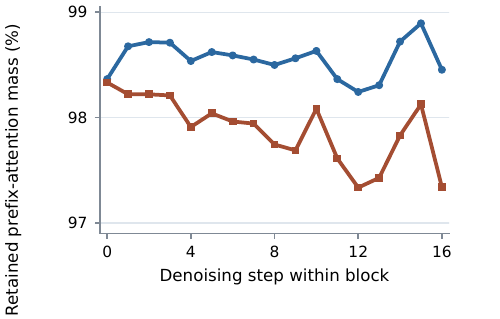}
\end{minipage}\hfill
\begin{minipage}[t]{0.47\linewidth}
\centering
\includegraphics[width=\linewidth]{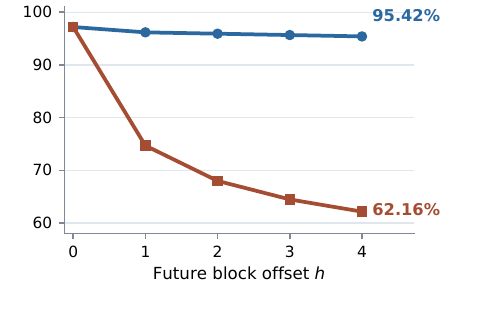}
\end{minipage}

\vspace{-4mm}
\par\smallskip
{\scriptsize
\textcolor{methodblue}{\rule[0.4ex]{1.2em}{0.6pt}}
Fresh selection at evaluation\qquad
\textcolor{methodrust}{\rule[0.4ex]{1.2em}{0.6pt}}
Reuse earlier selection
}
\par

\vspace{-4mm}
\begin{subfigure}[t]{0.47\linewidth}
\centering
\caption{Within a block}
\end{subfigure}\hfill
\begin{subfigure}[t]{0.47\linewidth}
\centering
\caption{Across blocks}
\end{subfigure}

\caption{Full-cache MATH-500 trace ($n=30$ initially). Within-block reuse stays close to step-specific selection; cross-block coverage declines as the prefix grows.}
\label{fig:attention-reuse}
\end{figure}

The horizons differ in a recorded full-cache MATH-500 trace (\Cref{fig:attention-reuse}).
At denoising step 8 of the current block, a shared read set covers 97.75\% of prefix-attention mass, against 98.50\% for a freshly selected set at the same budget ($n=29$ at that step).
One block later, coverage by the old set is 74.69\%, and after four blocks it is 62.16\%.
The fresh-set comparison includes entries appended after the old set was selected.
The trace therefore measures loss of coverage as the prefix grows, including newly available positions.
It motivates refreshing read selection each block but cannot, by itself, identify the best retention rule.

\subsection{Queries available before commitment}
The first denoising step of block $t$ contains $B$ masks, with queries $\mathcal Q_t^{(1)}$.
These queries are available before any token in the block is finalized and can rank reads for its remaining steps.
Retention needs evidence about blocks that have not yet been generated.
At each eviction checkpoint, a block-causal, no-write probe places masks at positions in the next $M-1$ blocks and returns $\mathcal Q_t^{(2)},\ldots,\mathcal Q_t^{(M)}$.
The probe does not append KV or produce draft tokens.
Its future-block queries depend on the known prefix and masks, since the intermediate blocks have not been finalized.
They estimate future cache demand rather than reproduce eventual queries.

We set the eviction interval to $E=4$ blocks and probe $M=4$ blocks, including the current one and the next three.
The retained cache also serves the first step of block $t+4$, which this probe does not cover. Current-block queries are included as available evidence, although the current block reads from its gathered buffer after eviction.
The horizon is therefore an empirical design choice, not an exact enumeration of the queries served by $S_t$. The fixed-budget controls in \Cref{tab:probe-ablation} compare it with shorter and longer probes.

\section{\method: Two Decisions, Two Query Horizons}
\label{sec:method}

\method builds two rankings from the queries available at block $t$.
The first ranking serves the remaining denoising steps of that block.
The second ranks entries kept in persistent KV until the next eviction.
This changes the evidence used for each decision while retaining one scoring rule.

\paragraph{A score for attention-output sensitivity.}
For a query $q$, let $p_{q,i}$ be the attention weight of live prefix entry $i$, normalized over the live prefix, and let $\boldsymbol v_i$ be its value.
Writing $\boldsymbol o_q=\sum_i p_{q,i}\boldsymbol v_i$, deletion of a single entry and renormalization give
\begin{equation}
\left\|\boldsymbol o_q-\boldsymbol o_{q,-i}\right\|_2
=\frac{p_{q,i}}{1-p_{q,i}}\left\|\boldsymbol v_i-\boldsymbol o_q\right\|_2.
\label{eq:single-removal}
\end{equation}
Thus entries with the same attention mass can have different effects on the output.
The identity applies to one deletion and does not solve the subset-selection problem.
For a query population $\mathcal Q$, we use
\begin{equation}
a_i(\mathcal Q)=\frac{1}{|\mathcal Q|}\sum_{q\in\mathcal Q}p_{q,i},\qquad
\boldsymbol c(\boldsymbol w)=\frac{\sum_jw_j\boldsymbol v_j}{\sum_jw_j},\qquad
f_i(\boldsymbol w)=w_i\|\boldsymbol v_i-\boldsymbol c(\boldsymbol w)\|_2.
\label{eq:ranking}
\end{equation}
The pooled score weights each value's deviation from the attention centroid by its relevance to the query population.
\method computes separate read and retention rankings with this multi-entry removal proxy.
The score ablation in \Cref{tab:factorized-main} checks its empirical contribution without treating the single-entry identity as a guarantee about answer accuracy.

\begin{figure}[t]
\centering
\includegraphics[width=0.9\linewidth]{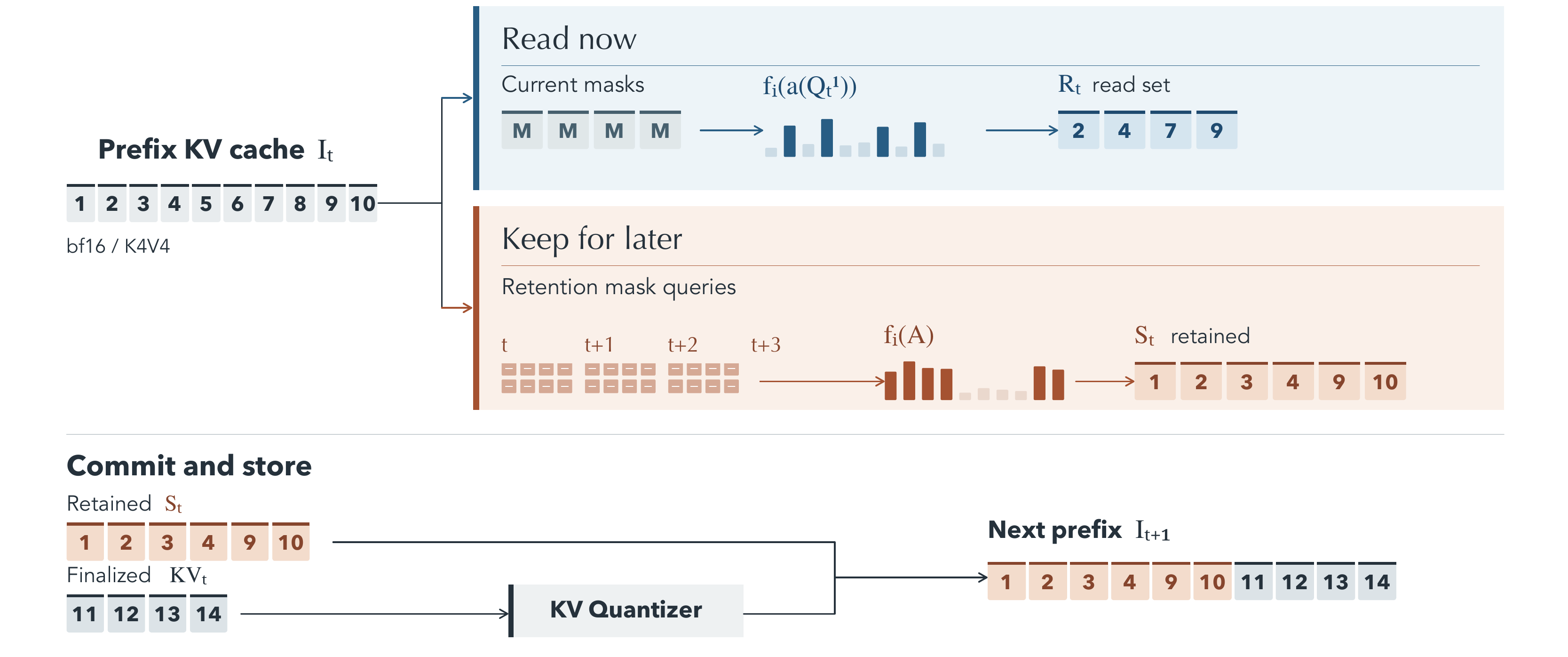}
\caption{Current-block masks select temporary reads. Current and future mask queries rank entries retained until the next eviction.}
\label{fig:two-horizons}
\end{figure}

\paragraph{Current masks select reads. Future masks guide retention.}
Let $\boldsymbol a(\mathcal Q)$ collect the weights $a_i(\mathcal Q)$ across live entries.
At the first step of block $t$, \method computes
\begin{equation}
s_i^{\rm read}=f_i\!\left(\boldsymbol a(\mathcal Q_t^{(1)})\right),\qquad
s_i^{\rm retain}=f_i\!\left(\sum\nolimits_{m=1}^{M}\boldsymbol a(\mathcal Q_t^{(m)})\right).
\label{eq:eviction-score}
\end{equation}
The top-$k_t$ read entries are gathered into a temporary buffer.
At an eviction checkpoint, four initial positions and a 64-position recent window are protected within the $C_t$ budget. The remaining slots go to the highest-scoring entries.
Ranking and budgets are applied per layer and KV head.
The first two layers remain dense for read selection.
The read buffer is gathered before persistent KV is compacted, so the current block still has its selected entries.
After its remaining denoising steps, the finalized block appends KV at its original logical positions.
These operations follow the order in \Cref{alg:maskahead}.

\begin{algorithm}{\linewidth}{\method uses separate query horizons for temporary reads and persistent KV retention.}
\label{alg:maskahead}
\normalfont\small\setlength{\parskip}{0pt}
\begin{tabularx}{\linewidth}{@{}r@{\hspace{.6em}}X@{}}
1 & \textbf{Input}: live cache $\mathcal I_t$, budgets $k_t,C_t$, horizon $M$, interval $E$. \\
2 & \textbf{Run} the first all-mask step of block $t$ against $\mathcal I_t$; collect $\mathcal Q_t^{(1)}$. \\
3 & \textbf{Select reads:} $R_t\leftarrow\operatorname{TopK}(s^{\rm read},k_t)$; \textbf{gather} $R_t$ into a temporary read buffer. \\
4 & \textbf{if} $t$ is an eviction checkpoint \textbf{then} \\
5 & \quad \textbf{Probe} masks in blocks $t+1,\ldots,t+M-1$ without writing persistent KV. \\
6 & \quad \textbf{Rank} the live prefix by $s^{\rm retain}$; protect sinks and recent positions. \\
7 & \quad \textbf{Compact} persistent KV to the retained set of budget $C_t$. \\
8 & \textbf{end if} \\
9 & \textbf{Denoise} the remaining steps using $R_t$; finalize block $t$ and append its KV. \\
\end{tabularx}
\end{algorithm}

\paragraph{Decisions from four-bit storage.}
\qmethod stores keys and values at four bits and computes rankings from that stored representation without a bf16 shadow cache.
Keys use asymmetric per-channel quantization over groups of 32 positions.
Values use per-token groups of 32 channels with fp16 metadata.
Incomplete key groups remain bf16, and the scoring pass unpacks stored values when needed.
After eviction, complete surviving key groups are copied bit for bit and partial groups are regrouped.
The four-bit result therefore tests selection, retention, and attention with a compact cache rather than quantizing a preselected bf16 cache.

For long prompts, the retention mechanism can be applied after full prefill or during prompt processing.
We compare full prefill, chunked prefill, and question-probe heuristics in \Cref{sec:prefill-results}.

\section{Experiments}
\label{sec:experiments}
We evaluate Fast-dLLM-v2-7B~\citep{wu2025fastdllmv2}, DreamReasoner-8B~\citep{wu2026dreamreasoner}, and LLaDA-2.0-mini~\citep{bie2025llada2}.

\paragraph{Tasks.}
Long-generation reasoning includes MATH-500 ($n=500$), containing competition-level mathematics problems, and GSM8K ($n=1319$), containing grade-school mathematical word problems, both scored by answer accuracy.
Long-prompt QA includes HotpotQA, NarrativeQA, and MuSiQue ($n=200$ each), covering multi-hop factual QA, narrative comprehension, and compositional multi-hop QA, respectively, scored by token F1.
Needle-in-a-haystack (NIAH) retrieval tests recovery of a target passage embedded in a long context ($n=200$), scored by retrieval accuracy.

\paragraph{Compared methods.}
We compare dense inference, sparse-read baselines, cache-eviction baselines, and our methods:
\begin{itemize}[nosep,leftmargin=*]    \item Dense: reads and retains the full prefix cache in bf16.
    \item MAGE~\citep{kwon2026mage}: uses attention from the fully masked first denoising step to select a prefix subset reused across the remaining steps. The full cache stays on the GPU.
    \item HERALD~\citep{kwon2026herald}: selects prefix entries using a representative mask query and retrieves them from a full cache offloaded to host memory.
    \item H2O~\citep{zhang2023h2o}: retains entries with high accumulated historical attention, together with protected initial and recent positions.
    \item H2O+CAOTE~\citep{zhang2023h2o,goel2025caote}: augments historical attention scores with value information to estimate the effect of eviction on attention outputs.
    \item Expected Attention~\citep{devoto2025expected}: estimates future attention from a query distribution fitted to past hidden states and retains entries with high expected importance.
    \item DapQ~\citep{tian2026dapq}: constructs surrogate future queries from pseudo tokens at future positions and uses them to rank cache entries for retention.
    \item \method: uses current-block mask queries for temporary read selection and future-block mask queries for persistent retention, with value-aware scoring and physical cache compaction.
    \item \qmethod: combines \method{} with 4-bit key/value storage to reduce the memory occupied by retained entries.
\end{itemize}

\paragraph{Protocol.}
Fast-dLLM-v2-7B is evaluated with greedy decoding.
For reasoning, the read budget is
$k_t=\min(N_t,\max(128,\lceil0.05N_t\rceil))$,
and the retention budget is
$C_t=\max(256,\lfloor0.1N_t^{\rm seen}\rfloor)$.
Here $N_t$ is the live prefix and $N_t^{\rm seen}$ counts positions seen before any eviction.
Long-context QA uses floors of 256 reads and 512 retained entries.
Baseline adaptations share the inference harness, decoding rule, per-task budgets, four protected initial positions, a 64-position recent window, and a four-block eviction interval where applicable.

KV memory includes retained keys and values and method state, normalized within each model and task to dense bf16.
For reasoning, we measure the maximum during the run. For QA and retrieval, we measure the resident cache after the response.
The prefill experiments report the maximum across prompt processing and generation (\Cref{sec:prefill-results}).
Bold and underlining rank the best and second distinct values, maximizing quality and minimizing KV, within the comparison stated by each table.
The separate batch-profile pilot reports absolute GiB and seconds (\Cref{sec:batch-profile}), rather than task-normalized KV ratios.

\begin{table*}[!b]
\centering
\caption{Reasoning accuracy (\%) / KV memory relative to dense bf16.}
\label{tab:eviction-main}
\small\renewcommand{\arraystretch}{1.03}\setlength{\tabcolsep}{5.5pt}
\begin{tabular}{@{}llccc@{}}
\toprule
& & \multicolumn{2}{c}{Fast-dLLM-v2-7B} & DreamReasoner-8B \\
\cmidrule(lr){3-4} \cmidrule(l){5-5}
Category & Method & MATH-500 & GSM8K & MATH-500 \\
\midrule
& Dense & 58.60 / 1.00 & 86.13 / 1.00 & 75.20 / 1.00 \\
\multirow{2}{*}{Selection}
& HERALD & 48.40 / 1.00 & 78.85 / 1.00 & 67.40 / 1.00 \\
& MAGE & 52.20 / 1.00 & 85.44 / 1.00 & 72.80 / 1.00 \\
\midrule
\multirow{4}{*}{Eviction}
& H2O & 50.80 / 0.62
      & 81.50 / $\second{0.85}$
      & 66.80 / $\second{0.68}$ \\
& H2O+CAOTE & 49.60 / $\second{0.61}$
            & $\best{85.44}$ / 1.01
            & 67.80 / $\second{0.68}$ \\
& Expected Attention & $\best{55.20}$ / 2.13
                     & $\second{84.91}$ / 3.03
                     & $\best{69.80}$ / 1.37 \\
& DapQ & 51.80 / 0.62
       & 83.85 / 0.86
       & 65.40 / $\second{0.68}$ \\
\rowcolor{csand!70}
& \method & $\second{54.60}$ / 0.62
          & $\second{84.91}$ / $\second{0.85}$
          & $\second{68.00}$ / $\second{0.68}$ \\
\rowcolor{csand!35}
\multirow{-2}{*}{\makecell[l]{Selection \&\\Eviction}}
& \qmethod & 54.00 / $\best{0.25}$
           & 84.23 / $\best{0.32}$
           & 67.40 / $\best{0.28}$ \\
\bottomrule
\end{tabular}
\end{table*}

\begin{table*}[t]
\centering
\caption{LongBench quality (F1) / KV relative to dense bf16.}
\label{tab:longbench-main}
\scriptsize\renewcommand{\arraystretch}{1.03}\setlength{\tabcolsep}{3.2pt}
\resizebox{\textwidth}{!}{%
\begin{tabular}{@{}ll*{6}{c}@{}}
\toprule
& & \multicolumn{3}{c}{Fast-dLLM-v2} & \multicolumn{3}{c}{LLaDA-2.0-mini} \\
\cmidrule(lr){3-5} \cmidrule(l){6-8}
Category & Method & HotpotQA & NarrativeQA & MuSiQue & HotpotQA & NarrativeQA & MuSiQue \\
\midrule
& Dense & 41.31 / 1.000 & 24.56 / 1.000 & 17.43 / 1.000
        & 59.82 / 1.000 & 26.93 / 1.000 & 40.46 / 1.000 \\
\multirow{2}{*}{Selection}
& HERALD & 38.02 / 1.000 & 24.97 / 1.000 & 17.13 / 1.000
         & 57.17 / 1.000 & 26.71 / 1.000 & 37.37 / 1.000 \\
& MAGE & 38.14 / 1.000 & 25.70 / 1.000 & 16.78 / 1.000
       & 57.17 / 1.000 & 27.03 / 1.000 & 39.49 / 1.000 \\
\midrule
\multirow{4}{*}{Eviction}
& H2O & 31.91 / 0.129 & 22.31 / 0.116 & 13.91 / 0.119
      & 48.32 / 0.128 & 23.72 / 0.115 & 32.25 / 0.119 \\
& H2O+CAOTE & 31.72 / 0.129 & 22.00 / 0.116 & 15.21 / 0.119
            & 48.18 / 0.128 & 24.53 / 0.115 & 31.48 / 0.119 \\
& Expected Attention & 22.52 / 0.209 & 21.81 / 0.163 & 9.28 / 0.175
                     & 40.98 / 0.173 & 21.25 / 0.141 & 21.21 / 0.151 \\
& DapQ & $\second{38.35}$ / $\second{0.107}$
       & $\second{24.86}$ / $\second{0.104}$
       & 16.74 / $\second{0.104}$
       & $\second{55.47}$ / $\second{0.107}$
       & $\best{27.59}$ / $\second{0.104}$
       & $\second{39.04}$ / $\second{0.104}$ \\
\rowcolor{csand!70}
& \method & $\best{38.41}$ / $\second{0.107}$
          & $\best{24.94}$ / $\second{0.104}$
          & $\best{17.15}$ / $\second{0.104}$
          & $\best{56.27}$ / $\second{0.107}$
          & $\best{27.59}$ / $\second{0.104}$
          & $\best{39.17}$ / $\second{0.104}$ \\
\rowcolor{csand!35}
\multirow{-2}{*}{\makecell[l]{Selection \&\\Eviction}}
& \qmethod & 34.75 / $\best{0.050}$
           & 23.90 / $\best{0.051}$
           & $\second{17.11}$ / $\best{0.048}$
           & 54.87 / $\best{0.053}$
           & $\second{27.24}$ / $\best{0.045}$
           & 38.99 / $\best{0.052}$ \\
\bottomrule
\end{tabular}}
\end{table*}

\subsection{Quality under a bounded cache}
\label{sec:quality-frontier}
Retention quality matters most when the cache cannot preserve many alternative candidates.
Accuracy does not increase monotonically with retention capacity at a fixed read budget, so storing more entries alone does not resolve all errors from sparse inference.

The main task matrix separates answer quality from the memory needed to obtain it (\Cref{tab:eviction-main,tab:longbench-main}).
On Fast-dLLM reasoning, \method retains much of the quality of full-cache sparse reads while physically discarding prefix entries.
Its advantage over full-cache MAGE on MATH-500 shows that cache capacity alone does not determine accuracy: the query source and read score also change between these policies.
Dense inference remains more accurate, and the relative ordering among evicting methods varies across tasks.

Expected Attention illustrates why the retained-token budget is insufficient for comparing memory.
Its Gaussian query state adds storage beyond the retained K/V, taking KV memory above the dense cache despite eviction.
Its strong MATH-500 accuracy therefore belongs to a different memory operating point.
The comparison must account for auxiliary state as well as retained entries, rather than interpret eviction itself as a guarantee of compression.

Long-prompt QA exposes a different pattern: \method and DapQ are close at matched KV memory, while both preserve substantially more quality than historical-attention retention.
\method has the highest evicting-policy F1 in all six model/task settings, including one tie.
Across these settings, the main result is an average $9.5\times$ reduction in KV memory with a mean 1.2-point F1 loss.

\qmethod further reduces the storage cost of surviving entries, giving the smallest KV footprint in every main model/task setting.
On reasoning tasks, this additional compression preserves accuracy close to bf16 \method.
Across long-prompt QA, it increases the average memory reduction to $20.1\times$ with a mean 2.3-point F1 loss relative to dense inference.
Thus, quantization extends the memory savings beyond eviction alone, with an additional quality cost that varies across tasks.

\subsection{What evidence does each decision need?}
\label{sec:mechanism-results}

\begin{wraptable}{r}{0.54\textwidth}
\vspace{-2mm}
\centering
\captionsetup{font=footnotesize,skip=3pt}
\caption{Probe controls on Fast-dLLM-v2-7B: MATH-500 ($n=500$) and HotpotQA ($n=200$). KV is measured on MATH-500 relative to dense bf16. $M$ counts the current and future probed blocks.}
\label{tab:probe-ablation}
\label{tab:mechanism-pair}
\footnotesize
\setlength{\tabcolsep}{3pt}
\begin{tabular}{@{}lrrr@{}}
\toprule
Probe & MATH (\%) & HotpotQA (F1) & KV \\
\midrule
$M=1$ & 52.0 & 38.26 & 0.625 \\
$M=2$ & $\second{54.4}$ & 37.86 & 0.625 \\
\rowcolor{csand!70}
$M=4$ & $\best{54.6}$ & $\best{38.41}$ & $\second{0.624}$ \\
$M=8$ & 53.2 & $\second{38.29}$ & $\best{0.621}$ \\
$M=4$, random tokens & 53.6 & $\second{38.29}$ & 0.625 \\
$M=4$, DapQ tokens & 51.2 & 38.01 & 0.625 \\
\bottomrule
\end{tabular}
\vspace{-2mm}
\end{wraptable}

Future-mask evidence improves retention without relying on a larger resident cache.
The query-source control fixes the score, budgets, and eviction interval, with nearly unchanged MATH-500 KV memory across variants (\Cref{tab:probe-ablation}).
Looking beyond the current block improves reasoning accuracy, but extending the probe further does not give a monotonic benefit.
Two and four blocks have similar point estimates, which favors a short lookahead over an unrestricted horizon.

\begin{wraptable}{r}{0.52\textwidth}
\centering
\captionsetup{font=footnotesize,skip=3pt}
\caption{Read and retention scores in bf16 at fixed queries and budgets: MATH-500 ($n=500$, $k=128,C=256$) and HotpotQA ($n=200$, $k=256,C=512$). Full denotes no eviction. KV is measured on MATH-500 relative to dense bf16.}
\label{tab:factorized-main}
\label{tab:factorization-two-tasks}
\footnotesize
\setlength{\tabcolsep}{3pt}
\begin{tabular}{@{}llrrr@{}}
\toprule
Read & Retain & MATH (\%) & HotpotQA (F1) & KV \\
\midrule
Mass & Full & 52.2 & $\second{38.14}$ & 1.000 \\
Value-aware & Full & $\second{53.8}$ & 37.85 & 1.000 \\
\midrule
Mass & Mass & 52.0 & 37.84 & 0.625 \\
Mass & Value-aware & $\second{53.8}$ & 38.08 & 0.625 \\
Value-aware & Mass & 53.4 & 37.91 & $\best{0.618}$ \\
\rowcolor{csand!70}
Value-aware & Value-aware & $\best{54.6}$ & $\best{38.41}$ & $\second{0.624}$ \\
\bottomrule
\end{tabular}
\vspace{-2mm}
\end{wraptable}

Masks provide better retention evidence than DapQ-filled future positions on MATH-500: the paired mask-versus-DapQ comparison has 28 wins and 11 losses ($p=0.0095$, two-sided sign test).
Unlike pseudo tokens, masks do not commit the probe to an unfinalized continuation. The control compares query sources without isolating the cause of the quality difference.
HotpotQA is much less sensitive to query source under its 32-token output protocol, so the reasoning result should not be generalized to short-answer QA.
We use four blocks as a short horizon on the same scale as the eviction interval, rather than as a uniquely optimal probe length.

The score factorization distinguishes better retention evidence from a better ranking criterion (\Cref{tab:factorized-main}).
Changing the read score affects what the current block sees, while changing the retention score affects what later blocks can still access.
Using the value-aware score for both decisions gives the highest point estimates among the evicting combinations, but its paired quality intervals against mass scoring include zero on both tasks (\Cref{app:factorization}).
The single-entry deletion identity in \Cref{eq:single-removal} motivates value-aware scoring by quantifying how removing an entry changes the attention output.
The quality experiments provide clearer evidence for using future-mask queries than for replacing attention-mass scoring with value-aware scoring.

\subsection{Precision changes the memory and quality budgets}
\label{sec:precision-results}
Eviction and quantization reduce different parts of the storage cost: the number of resident entries and the bytes used to represent each entry.
Their combination gives the smallest measured KV footprint in the main task matrix (\Cref{tab:quant-selection}).
On MATH-500, four-bit storage adds a $2.48\times$ KV-memory reduction over bf16 \method at a 0.6-point accuracy cost.
The reduction is smaller than the nominal ratio between bf16 and four-bit payloads because the reported KV also includes method state, not only compressed entries.

The quality cost of stored precision is task dependent.
Untransformed K4V4 loses more F1 on Fast-dLLM HotpotQA than on the other main QA settings, while the rotated-and-centered key variant recovers much of that gap at unchanged KV memory (\Cref{tab:key-transform}).
This comparison shows that the quantization representation matters within the same storage budget, although it does not separate the effects of rotation and centering.
Across the six main QA settings, \qmethod reduces KV memory by $20.1\times$ on average at a mean 2.3-point F1 loss relative to dense inference.

High ranking overlap is insufficient to certify quality preservation.
The recording-only diagnostic finds over 97\% overlap for both reads and retention, yet the generated answers remain sensitive to precision (\Cref{app:precision-overlap}).
Even unchanged sets use perturbed keys and values in attention, and set overlap does not measure the importance of the entries that differ.

\begin{table*}[t]
\centering
\caption{Selection, retention, and stored precision on Fast-dLLM-v2-7B. MATH and GSM8K report accuracy (\%), and HotpotQA reports F1. KV memory is relative to dense bf16.}
\label{tab:quant-selection}
\small\setlength{\tabcolsep}{3pt}
\begin{tabular}{@{}lllrrrrr@{}}
\toprule
Read score & Format & Retain & MATH & GSM8K & HotpotQA & MATH KV & HotpotQA KV \\
\midrule
Mass & bf16 & Full & 52.2 & $\best{85.44}$ & $\second{38.14}$ & 1.000 & 1.000 \\
Mass & K4V4 & Full & 52.0 & 84.61 & 36.07 & $\second{0.313}$ & 0.313 \\
Value-aware & bf16 & Full & 53.8 & $\second{84.91}$ & 37.85 & 1.000 & 1.000 \\
\rowcolor{csand!70}
Value-aware & bf16 & Evict & $\best{54.6}$ & $\second{84.91}$ & $\best{38.41}$ & 0.624 & $\second{0.107}$ \\
\rowcolor{csand!70}
Value-aware & K4V4 & Evict & $\second{54.0}$ & 84.23 & 34.75 & $\best{0.252}$ & $\best{0.050}$ \\
\bottomrule
\end{tabular}
\end{table*}

The budget sweeps expose a separate quality lever: more reads can improve answers without enlarging the persistent cache (\Cref{fig:pareto-sweeps}).
At fixed retention, increasing the MATH-500 read floor brings accuracy close to dense inference while KV memory remains near the same level.
HotpotQA also benefits from more reads, but the QA tasks have different read-fraction orderings (\Cref{tab:read-fraction}).
Thus, the storage budget need not dictate one read budget for every workload.
These within-policy controls distinguish errors caused by limited access to surviving entries from the consequences of deleting entries, without establishing a global frontier over all baselines.

\subsection{Batch-profile memory and execution time}
\label{sec:batch-profile}
The systems pilot separates smaller storage from faster execution (\Cref{fig:batch-profile}).
K4V4-only storage substantially reduces resident KV but does not reduce the recorded batch-32 decode time.
Quantization alone leaves the full logical prefix available to the attention procedure, whereas \method also reduces the entries gathered for repeated denoising and removes entries from persistent storage.

At batch 32, \method and \qmethod reduce KV memory by 85.1\% and 89.9\%, respectively.
Their recorded decode-time ratios relative to dense bf16 are $1.68\times$ and $1.64\times$, while complete-batch ratios are smaller at $1.23\times$ and $1.20\times$.
Most of the difference in reported median times lies in decoding, which is the stage targeted by repeated sparse reads.

Each policy processes the same 32 requests within each of three seeded prompt cohorts and generates 256 token positions per request, with EOS stopping disabled.
The median end-to-end output-token throughputs are 55.6 tokens/s for dense bf16, 68.2 for \method, and 66.5 for \qmethod.
Thus, the complete-batch ratios compare matched fixed-length outputs, not differences in generated length.
The profile is separate from the quality experiments and does not measure throughput with natural stopping or at matched quality.
\Cref{app:batch-profile} reports the complete timings and output-length accounting.

\begin{figure}[t]
\centering
\begin{subfigure}[t]{0.485\linewidth}
\centering
\includegraphics[width=\linewidth]{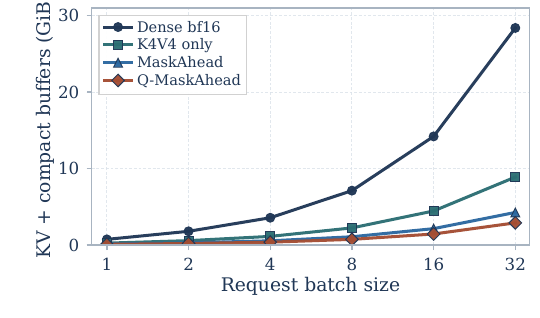}
\caption{KV memory}
\label{fig:batch-profile-kv}
\end{subfigure}\hfill
\begin{subfigure}[t]{0.485\linewidth}
\centering
\includegraphics[width=\linewidth]{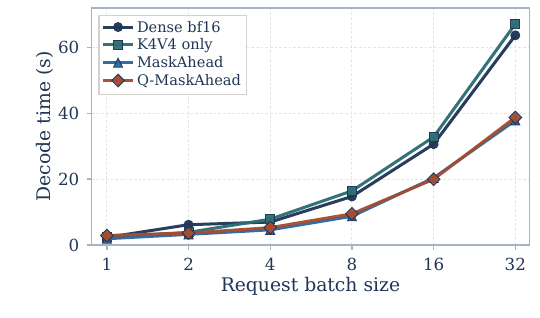}
\caption{Decode-stage time}
\label{fig:batch-profile-decode}
\end{subfigure}
\caption{Batch-profile medians over three prompt mixes, restricted to batches 1--32. Memory is measured after batch completion and includes resident KV and compact buffers. Decode timing excludes the other stages of the request. Complete-batch time and the batch-64 records are in \Cref{app:batch-profile}.}
\label{fig:batch-profile}
\end{figure}

\subsection{Retention heuristics for long prefill}
\label{sec:prefill-results}
The QA comparisons above use full prefill: the model processes the complete prompt before applying retention, once the question is available.
This reduces KV memory during decoding. When the prompt is much longer than the response, however, the memory peak occurs during prefill.
We therefore apply the same retention framework with three prefill heuristics:
\begin{itemize}[nosep,leftmargin=*]
    \item Full prefill: retain the prompt cache until prefill completes, then compress it for decoding.
    \item Chunked prefill: process the prompt in 512-token chunks and evict entries between chunks.
    \item Question probe: use question-conditioned probing to rank entries for early retention during prompt processing.
\end{itemize}

\begin{wraptable}{r}{0.3\textwidth}
\vspace{-2mm}
\centering
\captionsetup{font=footnotesize,skip=3pt}
\caption{\method{} prefill heuristics on HotpotQA ($n=200$, bf16).}
\label{tab:prefill}
\footnotesize
\setlength{\tabcolsep}{5pt}
\begin{tabular}{@{}lrr@{}}
\toprule
Heuristic & F1 & KV \\
\midrule
Full prefill & $\best{38.41}$ & 0.999 \\
Chunked prefill & 26.90 & $\second{0.185}$ \\
\rowcolor{csand!70}
Question probe & $\second{37.16}$ & $\best{0.144}$ \\
\bottomrule
\end{tabular}
\vspace{-2mm}
\end{wraptable}

Question probe preserves substantially more answer quality than chunked prefill while using less KV memory (\Cref{tab:prefill}).
Both apply retention before the prompt cache reaches its full size, but the question provides task-specific evidence for deciding which context to preserve.
The comparison changes both ranking and retention timing, so it evaluates the complete heuristics rather than isolating question conditioning.

Combining question probe with four-bit storage requires a separate quality check: this combination sharply lowers F1 on all three QA tasks (\Cref{tab:prefill-precision-boundary}).
NIAH retrieval provides a complementary test of retained information: bf16 \method and DapQ remain close to full-cache MAGE after compression, while four-bit storage incurs a larger retrieval loss (\Cref{tab:retrieval}).

\section{Conclusion}
\label{sec:conclusion}
\method combines current-block read selection with future-mask-guided retention in block diffusion.
It achieves the highest F1 among evicting methods across all six main QA settings, including one tie, with an average $9.5\times$ reduction in KV memory and a 1.2-point mean F1 loss relative to dense inference.
\qmethod extends the average reduction to $20.1\times$ with a 2.3-point F1 loss.

Matched controls favor future-block masks over current-only and pseudo-token probes.
Value-aware scoring has the highest point estimates in the score factorization, while paired quality intervals include zero.
The reasoning comparisons identify the budget dependence: the gap over H2O is largest at the smallest retention floor, while H2O+CAOTE leads at the wider MATH-500 budget.

\label{main:lastpage}
\end{mainpart}

\begin{appendixpart}
\numberwithin{table}{section}
\section{Batch Profiling and Cost Accounting}
\label{app:batch-profile}
The supplementary results follow the two cache decisions in the main text.
This section records the systems pilot and its memory metrics.
\Cref{app:budget-results} examines read and retention budgets, \Cref{app:mechanism-results} isolates query and score choices, \Cref{app:precision-results} studies stored precision, and \Cref{app:prefill-results} separates prefill from generation.

\subsection{Measurement scope and complete batch profile}
\label{app:profile-protocol}
The displayed profile contains four cache policies over batches 1--64. All plots stop at batch 32 so that timing differences remain visible.
The plotted medians aggregate three seeded prompt cohorts with the same requests across policies within each cohort.
The results record output-token throughput as generated positions divided by complete-batch wall time.
The source JSON retains per-cohort timing, token throughput, request IDs, prompt-token totals, and source/configuration hashes.
The fixed-output ratios do not establish throughput with natural stopping, fixed-quality throughput, or a serving-capacity limit.

Generation-end resident KV includes used entries and compact buffers.
The figures and tables use this KV metric. The source CSV also preserves a full-run device-allocation field, which is not used for the memory comparison here.
\Cref{fig:batch-profile-full} shows KV together with complete-batch and decode-stage times over the common successful range.
Only successful measurements have numerical coordinates.
At batch 64, the supplied summary records OOM for dense bf16 and K4V4-only storage. The completed \method run uses 8.5\,GiB of resident KV and takes 242.4\,s in total, including 78.6\,s for decode.
The \qmethod batch-64 run uses 5.8\,GiB and takes 245.0\,s in total, including 81.3\,s for decode.

\begin{figure}[!htbp]
\centering
\begin{subfigure}[t]{0.485\linewidth}
\centering
\includegraphics[width=\linewidth]{new_figs/batch_kv.pdf}
\caption{KV memory}
\end{subfigure}\hfill
\begin{subfigure}[t]{0.485\linewidth}
\centering
\includegraphics[width=\linewidth]{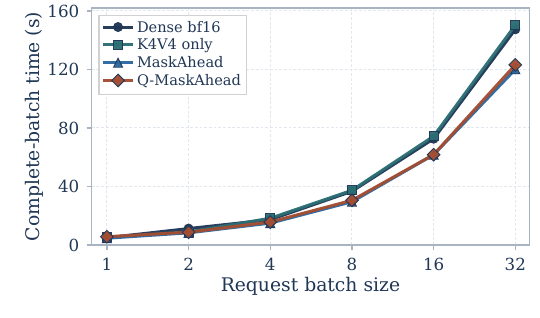}
\caption{Complete-batch time}
\end{subfigure}
\par\medskip
\begin{subfigure}[t]{0.485\linewidth}
\centering
\includegraphics[width=\linewidth]{new_figs/batch_decode.pdf}
\caption{Decode-stage time}
\end{subfigure}
\caption{Systems pilot over batches 1--32: resident KV, complete-batch time, and decode time. Times are supplied medians over three prompt mixes. Every plotted setting is successful. Memory is measured after batch completion and includes compact buffers. Batch-64 records are listed separately in \Cref{tab:batch-profile}.}
\label{fig:batch-profile-full}
\end{figure}
\FloatBarrier
\clearpage
\small\setlength{\tabcolsep}{6pt}
\begin{longtable}{@{}lrrrr@{}}
\caption{KV and timing profile. Memory is measured after batch completion in GiB, and time is in seconds. Values are rounded to one decimal place. OOM indicates the recorded run status.}\label{tab:batch-profile}\\
\toprule
Policy & Batch & KV & Wall time & Decode time \\
\midrule
\endfirsthead
\multicolumn{5}{l}{Table \thetable\ continued.}\\
\toprule
Policy & Batch & KV & Wall time & Decode time \\
\midrule
\endhead
\bottomrule
\endfoot
Dense bf16 & 1 & 0.7 & 4.9 & 2.3 \\
Dense bf16 & 2 & 1.8 & 11.1 & 6.1 \\
Dense bf16 & 4 & 3.5 & 17.2 & 6.9 \\
Dense bf16 & 8 & 7.1 & 36.6 & 14.7 \\
Dense bf16 & 16 & 14.2 & 72.6 & 30.6 \\
Dense bf16 & 32 & 28.4 & 147.4 & 63.7 \\
Dense bf16 & 64 & \multicolumn{3}{c}{OOM} \\
\midrule
K4V4 only & 1 & 0.2 & 5.1 & 2.4 \\
K4V4 only & 2 & 0.6 & 8.8 & 3.8 \\
K4V4 only & 4 & 1.1 & 18.2 & 7.8 \\
K4V4 only & 8 & 2.2 & 37.3 & 16.4 \\
K4V4 only & 16 & 4.4 & 74.3 & 32.8 \\
K4V4 only & 32 & 8.9 & 150.2 & 67.1 \\
K4V4 only & 64 & \multicolumn{3}{c}{OOM} \\
\midrule
\method & 1 & 0.1 & 4.5 & 1.9 \\
\method & 2 & 0.3 & 8.2 & 3.2 \\
\method & 4 & 0.5 & 14.9 & 4.6 \\
\method & 8 & 1.1 & 29.5 & 8.7 \\
\method & 16 & 2.1 & 61.6 & 20.3 \\
\method & 32 & 4.2 & 120.1 & 37.8 \\
\method & 64 & 8.5 & 242.4 & 78.6 \\
\midrule
\qmethod & 1 & 0.1 & 5.6 & 2.8 \\
\qmethod & 2 & 0.2 & 8.6 & 3.6 \\
\qmethod & 4 & 0.4 & 15.6 & 5.2 \\
\qmethod & 8 & 0.7 & 30.4 & 9.4 \\
\qmethod & 16 & 1.4 & 61.7 & 19.9 \\
\qmethod & 32 & 2.9 & 123.1 & 38.7 \\
\qmethod & 64 & 5.8 & 245.0 & 81.3 \\
\end{longtable}
\normalsize

\subsection{Generated length in the quality runs}
\label{app:generated-length}
The batch profile fixes the output length, whereas the quality evaluations use their configured stopping rules.
\Cref{tab:generated-length} reports mean generated tokens per example from the per-example runtime records for the Fast-dLLM-v2 configurations in the main quality tables.
All methods within a dataset are evaluated on the same examples.

\begin{table}[!htbp]
\centering
\caption{Mean generated tokens per example under the quality-evaluation stopping rules. The sample sizes are 500 for MATH-500, 1,319 for GSM8K, and 200 for each QA dataset. Method rows use the configurations reported in the main quality tables.}
\label{tab:generated-length}
\scriptsize\setlength{\tabcolsep}{3.5pt}
\begin{tabular}{@{}lrrrrr@{}}
\toprule
Method & MATH-500 & GSM8K & HotpotQA & NarrativeQA & MuSiQue \\
\midrule
Dense bf16 & 611.8 & 311.8 & 8.75 & 13.69 & 11.40 \\
HERALD & 605.0 & 310.8 & 9.25 & 14.11 & 11.42 \\
MAGE & 624.7 & 314.2 & 9.32 & 13.50 & 11.28 \\
H2O & 662.8 & 313.8 & 9.88 & 14.09 & 12.15 \\
H2O+CAOTE & 676.6 & 313.9 & 9.77 & 13.46 & 11.61 \\
Expected Attention & 652.4 & 312.3 & 11.14 & 13.91 & 13.25 \\
DapQ & 647.3 & 311.3 & 9.28 & 13.63 & 11.40 \\
\method & 617.8 & 313.6 & 9.28 & 13.64 & 11.36 \\
\qmethod & 646.8 & 313.7 & 10.04 & 15.26 & 11.83 \\
\bottomrule
\end{tabular}
\end{table}

On MATH-500, \method generates 617.8 tokens per example, 1.0\% more than dense bf16, while \qmethod generates 646.8, 5.7\% more.
The corresponding GSM8K increases are 0.6\% for both variants.
For the short-answer QA tasks, the bf16 variant differs from dense by at most 0.53 token per example.
The four-bit variant generates up to 1.6 additional tokens per example on NarrativeQA.
Thus, response length is not identical in the quality runs, especially for four-bit storage, and their latency cannot be inferred from the fixed-length batch profile.

\subsection{Amortized attention traffic}
\label{app:cost-model}
The pilot above measures execution time. The following accounting describes the attention-read component rather than predicting throughput.
Let $T$ be denoising steps per block, $N$ live prefix entries, and $k$ the temporary read budget.
Express probe and compaction work per checkpoint as $Q_{\rm probe}$ and $Q_{\rm compact}$ in equivalent prefix-entry reads.
Dense attention reads approximately $TN$ entries per block.
\method reads $N+(T-1)k$ for block attention, with $(Q_{\rm probe}+Q_{\rm compact})/E$ amortized over checkpoints.
This simplified traffic decreases when
\begin{equation}
(T-1)(N-k)>\frac{Q_{\rm probe}+Q_{\rm compact}}{E}.
\label{eq:traffic-break-even}
\end{equation}
The expression omits value-aware score computation, read-buffer gathering, dense-read exceptions, kernel launches, and request scheduling.
Resident-cache reduction therefore does not imply an equal reduction in total traffic or full-run memory allocation.

\FloatBarrier
\section{Read and Retention Budget Tradeoffs}
\label{app:budget-results}
Across the six model/task settings in \Cref{tab:longbench-main}, \method reduces KV memory by $9.5\times$ on average with a mean 1.2-point F1 loss relative to dense inference.
\qmethod increases the average reduction to $20.1\times$ with a mean 2.3-point loss.
The mean F1 difference between bf16 \method and \qmethod is 1.1 points in favor of bf16.
We average within-setting dense-to-method memory ratios and F1 differences.

The read budget controls the prefix gathered for repeated denoising, while the retention budget controls entries available to later blocks.
We first vary retention on MATH-500, then compare policies at a wider budget, and finally vary read and retention fractions on long-context QA.
These series use different floors, which are stated separately rather than pooled.

\subsection{Reasoning capacity sweep}
\label{app:math-capacity}
With the 128-entry read floor fixed, \Cref{tab:math-capacity} compares \method and H2O across four retention floors on the same 500 MATH-500 examples.
The accuracy difference is 11.6 points at $C=128$, 3.8 points at $C=256$, and 0.2 points at both $C=512$ and $C=1024$.
KV includes growth between four-block eviction checkpoints.
The large difference at the smallest floor is a comparison with H2O, while the broader baseline comparison uses the main $C=256$ floor.

\begin{table}[!htbp]
\centering
\caption{MATH-500 retention-floor sweep at fixed $k_{\min}=128$, bf16, on the same 500 examples. KV includes method state and is relative to dense bf16. Bold and underlining compare the two policies within each floor.}
\label{tab:math-capacity}
\small\setlength{\tabcolsep}{5pt}
\begin{tabular}{@{}rrrrr@{}}
\toprule
$C$ floor & \multicolumn{2}{c}{\method} & \multicolumn{2}{c}{H2O} \\
& Acc. (\%) & KV & Acc. (\%) & KV \\
\midrule
128 & $\best{51.0}$ & $\second{0.446}$ & $\second{39.4}$ & $\best{0.429}$ \\
256 & $\best{54.6}$ & $\second{0.624}$ & $\second{50.8}$ & $\best{0.617}$ \\
512 & $\best{53.0}$ & $\best{0.852}$ & $\second{52.8}$ & $\second{0.853}$ \\
1024 & $\best{53.8}$ & $\best{0.982}$ & $\second{53.6}$ & $\second{0.984}$ \\
\bottomrule
\end{tabular}
\end{table}

\FloatBarrier
\subsection{Common-harness comparison at a wider budget}
\label{app:wider-provenance}
The wider-budget runs use $k_t=\min(N_t,\max(256,\lceil0.1N_t\rceil))$ and $C_t=\max(512,\lfloor0.20N_t^{\rm seen}\rfloor)$, with eviction every four blocks.
Every displayed MATH-500 row has 500 examples and each QA row has 200.
All policies use the same examples and common inference harness, with bf16 storage except for \qmethod.
At this budget, historical CAOTE-style retention reaches 57.6\% on MATH-500, \method and Expected Attention reach 57.2\%, and DapQ reaches 56.2\% (\Cref{tab:wider-budget}).
Expected Attention uses 2.346 times dense bf16 KV memory because its Gaussian state dominates the retained cache.
\method has the highest F1 on HotpotQA and NarrativeQA, scoring 39.10 and 24.97 compared with 39.04 and 24.43 for DapQ. On MuSiQue, H2O leads at 18.02 F1, followed by \method at 17.17.
On GSM8K, H2O, DapQ, and \method each score 85.44\%, while \qmethod scores 86.35\% across 1,319 examples.
These complete greedy runs locate the budget regime in which retention rules differ.

\begin{table}[!htbp]
\centering
\caption{Fast-dLLM-v2-7B at $k=10\%$, $C=20\%$, with read/retention floors of 256/512. MATH-500 uses 500 examples and each QA task uses 200. Bold and underlining rank the evicting policies within each task.}
\label{tab:wider-budget}
\small\setlength{\tabcolsep}{8pt}
\begin{tabular}{@{}lrrrr@{}}
\toprule
Policy & MATH-500 & HotpotQA & NarrativeQA & MuSiQue \\
\midrule
H2O, bf16 & 56.4 & 33.09 & 24.36 & $\best{18.02}$ \\
H2O+CAOTE, bf16 & $\best{57.6}$ & 34.21 & 24.23 & 16.82 \\
DapQ, bf16 & 56.2 & $\second{39.04}$ & $\second{24.43}$ & 16.89 \\
Expected Attention, bf16 & $\second{57.2}$ & 24.94 & 23.79 & 11.99 \\
\rowcolor{csand!70}
\method & $\second{57.2}$ & $\best{39.10}$ & $\best{24.97}$ & $\second{17.17}$ \\
\rowcolor{csand!35}
\qmethod & 55.4 & 36.44 & 24.32 & 16.92 \\
\bottomrule
\end{tabular}
\end{table}

\FloatBarrier
\subsection{Read-budget controls at fixed retention}
\Cref{fig:pareto-sweeps} makes the read-budget dependence explicit while holding retention fixed.
On MATH-500, increasing the read floor from 128 to 512 raises accuracy from 54.6\% to 58.4\% while KV memory stays near 0.62 of dense bf16.
On HotpotQA, both bf16 and four-bit storage improve as the read fraction increases, while each curve retains its own fixed storage format and retention budget.
Retention-floor and precision comparisons are reported separately in \Cref{tab:math-capacity,tab:precision-grid}.
\begin{figure}[htbp]
\centering
\begin{subfigure}[t]{0.485\linewidth}
\centering
\includegraphics[width=\linewidth]{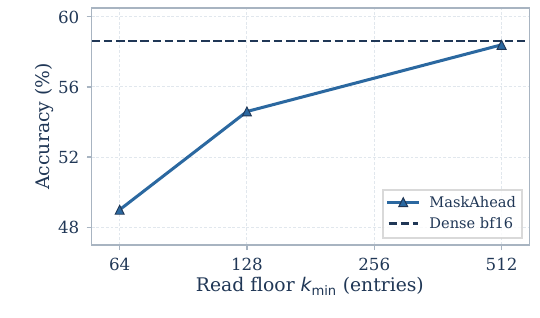}
\caption{MATH-500}
\end{subfigure}\hfill
\begin{subfigure}[t]{0.485\linewidth}
\centering
\includegraphics[width=\linewidth]{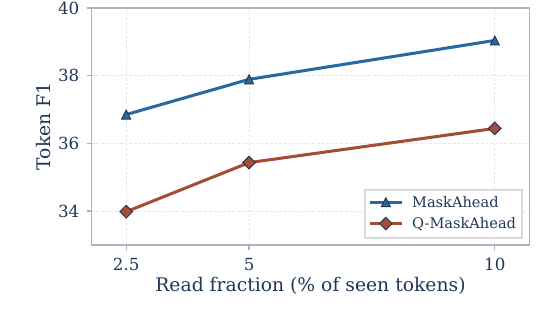}
\caption{HotpotQA}
\end{subfigure}
\caption{Read-budget controls on Fast-dLLM-v2-7B, with fixed retention: MATH-500 ($n=500$, $C_{\min}=256$) and HotpotQA ($n=200$, $C=20\%$). Only the read floor or fraction varies within each curve. The dashed line is dense bf16 MATH accuracy.}
\label{fig:pareto-sweeps}
\end{figure}

\FloatBarrier
\subsection{Long-context budget and read-fraction series}
\label{app:long-budget}
The long-context budget series uses $k=C/2$, expressed as fractions of seen tokens, with floors of 32 reads and 64 retained entries.
These floors differ from the main QA setting of 256 reads and 512 retained entries and from the wider-budget comparison above.
\method improves HotpotQA and NarrativeQA over H2O at all three budget fractions while using less KV memory (\Cref{tab:long-budget}). H2O leads MuSiQue by 0.30 F1 at $C=20\%$.

\begin{table}[!htbp]
\centering
\caption{LongBench retention-budget comparison of H2O and \method on Fast-dLLM-v2-7B, $n=200$ per task. $C$ is a fraction of seen tokens and $k=C/2$, with floors of 64 retained entries and 32 reads. KV is relative to dense bf16. These runs use full prefill, whose memory peak remains near dense bf16.}
\label{tab:long-budget}
\scriptsize\setlength{\tabcolsep}{4pt}
\begin{tabular}{@{}rlrrrrrr@{}}
\toprule
& & \multicolumn{2}{c}{HotpotQA} & \multicolumn{2}{c}{NarrativeQA} & \multicolumn{2}{c}{MuSiQue} \\
$C$ (\%) & Policy & F1 & KV & F1 & KV & F1 & KV \\
\midrule
5 & H2O & $\second{29.71}$ & $\second{0.077}$ & $\second{22.28}$ & $\second{0.066}$ & $\second{14.41}$ & $\second{0.069}$ \\
5 & \method & $\best{37.10}$ & $\best{0.054}$ & $\best{25.19}$ & $\best{0.053}$ & $\best{15.99}$ & $\best{0.053}$ \\
\midrule
10 & H2O & $\second{32.98}$ & $\second{0.127}$ & $\second{22.84}$ & $\second{0.116}$ & $\second{14.17}$ & $\second{0.119}$ \\
10 & \method & $\best{38.29}$ & $\best{0.105}$ & $\best{25.02}$ & $\best{0.104}$ & $\best{16.74}$ & $\best{0.104}$ \\
\midrule
20 & H2O & $\second{34.17}$ & $\second{0.226}$ & $\second{23.59}$ & $\second{0.216}$ & $\best{17.19}$ & $\second{0.219}$ \\
20 & \method & $\best{39.04}$ & $\best{0.206}$ & $\best{24.78}$ & $\best{0.206}$ & $\second{16.89}$ & $\best{0.206}$ \\
\bottomrule
\end{tabular}
\end{table}

At fixed $C=20\%$, increasing reads from 2.5\% to 10\% raises HotpotQA F1 from 36.85 to 39.04 at the same 0.206 KV memory.
NarrativeQA and MuSiQue have different read-fraction orderings (\Cref{tab:read-fraction}), so a single task does not determine the read budget for all QA workloads.

\begin{table}[!htbp]
\centering
\caption{Read-budget sweep at fixed $C=20\%$, with 200 examples per task. Retained-cache size is fixed within each storage format. Bold and underlining rank read fractions within each task and format.}
\label{tab:read-fraction}
\small\setlength{\tabcolsep}{4pt}
\begin{tabular}{@{}lrrrr@{}}
\toprule
KV & $k$ (\%) & HotpotQA F1 & NarrativeQA F1 & MuSiQue F1 \\
\midrule
bf16 & 2.5 & 36.85 & $\best{25.50}$ & 16.74 \\
bf16 & 5 & $\second{37.89}$ & $\second{24.96}$ & $\best{17.06}$ \\
bf16 & 10 & $\best{39.04}$ & 24.78 & $\second{16.89}$ \\
\midrule
K4V4 & 2.5 & 33.98 & $\second{24.23}$ & $\best{16.94}$ \\
K4V4 & 5 & $\second{35.43}$ & 23.68 & 16.83 \\
K4V4 & 10 & $\best{36.44}$ & $\best{24.39}$ & $\second{16.92}$ \\
\bottomrule
\end{tabular}
\end{table}

\FloatBarrier
\section{Query Horizons and Ranking Criteria}
\label{app:mechanism-results}
This section separates three kinds of evidence: attention-coverage diagnostics, controls over probe queries, and controls over the read/retention score.
Coverage uses recorded dense-cache trajectories. The other controls generate complete answers with matched cache budgets.

\subsection{Recorded within-block coverage}
\label{app:decision-horizon}
We compare a block-shared read set against a same-budget offline top-$k$ set selected from each later step's realized queries.
The full-cache MATH-500 recording starts with 30 examples, with fewer reaching later denoising steps.
The offline set is an attention-coverage reference and was not used to generate answers.

\begin{table}[!htbp]
\centering
\caption{Realized prefix-attention mass covered by a block-shared read set and same-budget step-specific offline top-$k$. The final column counts examples reaching each step, so rows have different sample sizes.}
\label{tab:step-coverage}
\small\setlength{\tabcolsep}{8pt}
\begin{tabular}{@{}rrrr@{}}
\toprule
Step & Shared (\%) & Offline top-$k$ (\%) & Examples \\
\midrule
0 & 98.33 & 98.37 & 30 \\
1 & 98.22 & 98.68 & 30 \\
4 & 97.91 & 98.54 & 30 \\
8 & 97.75 & 98.50 & 29 \\
12 & 97.34 & 98.24 & 24 \\
16 & 97.34 & 98.45 & 21 \\
\bottomrule
\end{tabular}
\end{table}

The shared set remains close to the offline reference within a block.
Across blocks, its overlap with the fresh set falls to 47.2\% at offset four, partly because the candidate prefix expands.
These diagnostics motivate refreshing reads. They do not compare retention-policy answer quality.

\FloatBarrier
\subsection{Probe horizon and query-source controls}
\label{app:probe-ablations}
We hold the read and retention budgets, value-aware score, and four-block eviction interval fixed while varying the probe horizon $M$.
The same future positions are also filled with random tokens or DapQ-style pseudo tokens.
The complete matched control is in \Cref{tab:probe-ablation} in the main text. It combines task quality with the corresponding MATH-500 KV memory measurements.

The masks-versus-DapQ comparison has 28 wins and 11 losses on paired MATH-500 examples, giving a two-sided sign-test $p=0.0095$.
Two-block and four-block masks have similar point estimates, 54.4\% and 54.6\%, while an eight-block probe scores 53.2\%.
The sweep supports a short lookahead rather than a monotonic benefit from more future positions.
On HotpotQA, all six variants lie within 0.55 F1 points under the 32-token output protocol.

\subsection{Independent read and retention scores}
\label{app:factorization}
The factorization in \Cref{tab:factorized-main} changes the read and retention scores independently while keeping query sources and budgets fixed.
Value-aware ranking for both decisions has the highest evicting-policy point estimate on MATH-500 and HotpotQA.
Relative to attention mass for both decisions, the differences are $+2.6$ MATH-500 points (paired 95\% bootstrap CI $[-0.6,5.8]$) and $+0.568$ HotpotQA F1 (CI $[-0.008,1.348]$).
Both intervals include zero.
Bootstrap samples are formed from aligned example IDs.

The single-entry deletion identity motivates the score, while these intervals leave its answer-quality contribution unresolved.
The probe controls above isolate query source with the scoring rule fixed.

\FloatBarrier
\Needspace{12\baselineskip}
\subsection{Protected positions and eviction interval}
The default policy protects four sinks and a recent window $W=64$, then evicts every $E=4$ blocks.

\begin{table}[!htbp]
\centering
\caption{Timing and protection controls on MATH-500 ($n=500$) at the main budgets. KV includes method state and is relative to dense bf16.}
\label{tab:timing-ablation}
\small\setlength{\tabcolsep}{5pt}
\begin{tabular}{@{}lrr@{}}
\toprule
Policy & MATH acc. (\%) & MATH KV \\
\midrule
\rowcolor{csand!70}
Default: sinks 4, $W=64$, $E=4$ & $\best{54.6}$ & 0.624 \\
No protected sinks & 53.0 & $\second{0.615}$ \\
$W=32$ & $\second{54.0}$ & 0.621 \\
$W=128$ & 53.6 & 0.618 \\
$E=2$ & 51.8 & $\best{0.543}$ \\
$E=8$ & 53.8 & 0.718 \\
\bottomrule
\end{tabular}
\end{table}
\Cref{tab:timing-ablation} changes these choices at the main budgets.
Evicting every two blocks lowers MATH-500 KV memory to 0.543 and accuracy to 51.8\%.
Evicting every eight blocks raises KV memory to 0.718 and scores 53.8\%.

\FloatBarrier
\section{Four-Bit Storage and Ranking Stability}
\label{app:precision-results}
Stored precision changes both the bytes per entry and the keys/values used for ranking and attention.
We examine key/value precision separately, then retention floors under K4V4, a key transform, and ranking overlap against bf16.
Early-retention quantization is evaluated with the prefill policies in \Cref{app:prefill-precision}.

\subsection{Key/value precision and retention floors}
\label{app:precision-grid}
At the main MATH-500 retention policy, K4V4 gives 54.0\% accuracy at 0.252 dense-relative KV memory.
Reducing key precision to two bits costs more accuracy than reducing value precision at similar cache size (\Cref{tab:precision-grid}).
The lowest-memory format is therefore a different accuracy operating point.

\begin{table}[!htbp]
\centering

\begin{minipage}[t]{0.48\linewidth}
\vspace{0pt}
\centering
\caption{MATH-500 KV precision under the same retention policy. KV is relative to dense bf16.}
\label{tab:precision-grid}
\scriptsize
\setlength{\tabcolsep}{3pt}
\begin{tabular}{@{}lrr@{}}
\toprule
KV format & Acc. & KV \\
\midrule
bf16 & $\best{54.6}$ & 0.624 \\
\rowcolor{csand!70}
K4V4 & $\second{54.0}$ & 0.252 \\
K4V2 & 53.4 & 0.203 \\
K2V4 & 48.2 & $\second{0.193}$ \\
K2V2 & 46.0 & $\best{0.148}$ \\
\bottomrule
\end{tabular}
\end{minipage}
\hfill
\begin{minipage}[t]{0.48\linewidth}
\vspace{0pt}
\centering
\caption{K4V4 MATH-500 retention floors at fixed $k=128$. KV is relative to dense bf16.}
\label{tab:fixed-k-budget}
\scriptsize
\setlength{\tabcolsep}{3pt}
\begin{tabular}{@{}rrr@{}}
\toprule
$C$ floor & Acc. & KV \\
\midrule
128 & 50.0 & $\best{0.156}$ \\
\rowcolor{csand!70}
256 & $\second{54.0}$ & $\second{0.252}$ \\
512 & $\best{54.2}$ & 0.336 \\
1024 & $\second{54.0}$ & 0.338 \\
\bottomrule
\end{tabular}
\end{minipage}

\end{table}

\label{app:four-bit-floors}
With $k_{\min}=128$ fixed, raising the K4V4 retention floor from 128 to 256 improves MATH-500 accuracy from 50.0\% to 54.0\%.
Larger floors add KV memory with little change in the measured accuracy (\Cref{tab:fixed-k-budget}).

\FloatBarrier
\subsection{Rotated and centered keys}
\label{app:key-transform}
At the same HotpotQA main budgets and 0.050 KV memory, the rotated-and-centered variant reaches 37.33 F1, compared with 34.75 for untransformed K4V4 (\Cref{tab:key-transform}).
These completed aggregates use 200 examples each.
Rotation and centering change together, so the comparison does not isolate their individual effects.
Per-example outputs for this pair are not in the supplied bundle and no paired interval is assigned.

\begin{table}[!htbp]
\centering
\caption{HotpotQA four-bit key variants at matched budgets. KV memory is normalized to dense bf16.}
\label{tab:key-transform}
\small
\begin{tabular}{@{}lrr@{}}
\toprule
Variant & Token F1 & KV \\
\midrule
Untransformed K4V4 & 34.75 & 0.050 \\
Rotated and centered keys & 37.33 & 0.050 \\
\bottomrule
\end{tabular}
\end{table}

\FloatBarrier
\subsection{Ranking overlap under stored precision}
\label{app:precision-overlap}
In a recording-only diagnostic, the same selection and retention scores are recomputed from bf16 and K4V4 on 30 MATH-500 and 30 HotpotQA examples.
The selected top-$k$ sets overlap by 97.44\% on average over 60 examples and 64,896 layer/head events, with example-level bootstrap interval $[97.12,97.77]\%$.
The retained top-$C$ sets overlap by 97.65\% over 48 examples with an eviction event and 9,968 layer/head events, with interval $[97.42,97.90]\%$.
High set overlap measures rank stability. It does not establish equal attention outputs or answer quality, as the HotpotQA K4V4 quality loss shows.

\FloatBarrier
\section{Prefill Policies and Long-Prompt Retrieval}
\label{app:prefill-results}
We compare full prefill, chunked prefill, and question probe, followed by their interaction with four-bit storage and passage retrieval.
Unless stated otherwise, the prefill comparisons measure the maximum KV memory over prompt processing and generation.

\subsection{Prefill heuristics}
Full prefill applies retention after processing the complete prompt.
On HotpotQA, it reaches 38.41 F1 with a maximum KV memory of 0.999 relative to dense bf16.
Question probe uses question-conditioned evidence for early retention and reaches 37.16 F1 at 0.144 KV memory.
Chunked prefill evicts between 512-token chunks and reaches 26.90 F1 at 0.185 KV memory.
All three leave the same normalized resident cache of 0.107 after the response, despite their different memory requirements during prompt processing.
These heuristics change ranking and retention timing together.
Question probe also reaches 24.10 F1 at 0.125 KV memory on NarrativeQA and 15.35 F1 at 0.129 on MuSiQue.

\FloatBarrier
\subsection{Precision boundary for early retention}
\label{app:prefill-precision}
Applying K4V4 at the same early-retention timing lowers KV memory further but also lowers F1 on all three QA tasks (\Cref{tab:prefill-precision-boundary}).
The bf16 prefill result and four-bit generation-time storage are separate operating points.
Their combination requires its own quality evaluation.

\begin{table}[!htbp]
\centering

\begin{minipage}[t]{0.48\linewidth}
\vspace{0pt}
\centering
\caption{Question-conditioned prefill with bf16 and K4V4 ($n=200$ per task), using the same early-retention timing. F1 is token F1; KV is relative to dense bf16.}
\label{tab:prefill-precision-boundary}
\scriptsize
\setlength{\tabcolsep}{3pt}
\begin{tabular}{@{}llrr@{}}
\toprule
Task & KV format & F1 & KV \\
\midrule
HotpotQA & bf16 & 37.16 & 0.144 \\
 & K4V4 & 10.79 & 0.073 \\
NarrativeQA & bf16 & 24.10 & 0.125 \\
 & K4V4 & 5.70 & 0.067 \\
MuSiQue & bf16 & 15.35 & 0.129 \\
 & K4V4 & 6.74 & 0.069 \\
\bottomrule
\end{tabular}
\end{minipage}
\hfill
\begin{minipage}[t]{0.48\linewidth}
\vspace{0pt}
\centering
\caption{NIAH passage retrieval ($n=200$). Accuracy is percent; maximum and resident KV are relative to dense bf16. Resident KV is measured after the response. Bold and underlining rank evicting policies.}
\label{tab:retrieval}
\scriptsize
\setlength{\tabcolsep}{2.5pt}
\begin{tabular}{@{}lrrr@{}}
\toprule
Policy & Acc. & Max. KV & Resident KV \\
\midrule
Dense & 83.0 & 1.0000 & 1.0000 \\
MAGE & 83.5 & 1.0000 & 1.0000 \\
H2O & 66.0 & 0.9999 & 0.1225 \\
H2O+CAOTE & 68.0 & 0.9999 & 0.1225 \\
Expected Attention & 21.5 & 1.0702 & 0.1929 \\
DapQ & $\second{82.0}$ & $\second{0.9995}$ & $\second{0.1042}$ \\
\rowcolor{csand!70}
\method & $\best{82.5}$ & $\second{0.9995}$ & $\second{0.1042}$ \\
\rowcolor{csand!70}
\qmethod & 75.5 & $\best{0.3137}$ & $\best{0.0503}$ \\
\bottomrule
\end{tabular}
\end{minipage}

\end{table}

\subsection{NIAH passage retrieval}
\label{app:official-retrieval}
On 200 NIAH passage-retrieval examples, DapQ scores 82.0\% and \method scores 82.5\% at the same 0.1042 KV memory.
Full-cache MAGE scores 83.5\%.
\qmethod scores 75.5\% at 0.0503 KV memory (\Cref{tab:retrieval}).
This comparison measures retrieval under compressed storage, with full bf16 prefill and its memory peak near dense.
\end{appendixpart}
\end{document}